\documentclass{article}

\usepackage[preprint]{neurips_2025}

\usepackage[utf8]{inputenc} %
\usepackage[T1]{fontenc}    %
\usepackage{hyperref}       %
\usepackage{url}            %
\usepackage{booktabs}       %
\usepackage{amsfonts}       %
\usepackage[table]{xcolor}  %
\usepackage{graphicx}       %
\usepackage{nicefrac}       %
\usepackage{microtype}      %
\usepackage{xcolor}         %
\usepackage{algorithm}
\usepackage{algorithmic}
\usepackage{amsmath}
\usepackage{cleveref} 
\newcommand{\redContent}[1]{{\color{red}#1}}
\newcommand{\blueContent}[1]{{\color{blue}#1}}

\usepackage{makecell}
\usepackage{xspace}
\usepackage{wrapfig}
\usepackage{caption}

\usepackage{float}  %

\usepackage{algorithm}  %
\usepackage{algorithmic}%

\title{DCI: Dual-Conditional Inversion for Boosting Diffusion-Based Image Editing}

\author{
  Zixiang Li\textsuperscript{1,2}, 
  Haoyu Wang\textsuperscript{1,2}, 
  Wei Wang\textsuperscript{1,2}, 
  Chuangchuang Tan\textsuperscript{1,2}, 
  Yunchao Wei\textsuperscript{1,2},
  Yao Zhao\textsuperscript{1,2} \\ 
 \textsuperscript{1}Institute of Information Science, Beijing Jiaotong University \\
  \textsuperscript{2}Visual Intelligence +X International Cooperation Joint Laboratory of MOE \\
}

\usepackage{caption}
\begin{document}

\maketitle

\begin{abstract}
Diffusion models have achieved remarkable success in image generation and editing tasks. Inversion within these models aims to recover the latent noise representation for a real or generated image, enabling reconstruction, editing, and other downstream tasks. However, to date, most inversion approaches suffer from an intrinsic trade-off between reconstruction accuracy and editing flexibility. This limitation arises from the difficulty of maintaining both semantic alignment and structural consistency during the inversion process. In this work, we introduce \textbf{Dual-Conditional Inversion (DCI)}, a novel framework that jointly conditions on the source prompt and reference image to guide the inversion process. Specifically, DCI formulates the inversion process as a dual-condition fixed-point optimization problem, minimizing both the latent noise gap and the reconstruction error under the joint guidance. This design anchors the inversion trajectory in both semantic and visual space, leading to more accurate and editable latent representations. Our novel setup brings new understanding to the inversion process. Extensive experiments demonstrate that DCI achieves state-of-the-art performance across multiple editing tasks, significantly improving both reconstruction quality and editing precision. Furthermore, we also demonstrate that our method achieves strong results in reconstruction tasks, implying a degree of robustness and generalizability approaching the ultimate goal of the inversion process.

\end{abstract}

\section{Introduction}
\begin{figure}[t]
    \centering
    \includegraphics[width=1.0\linewidth]{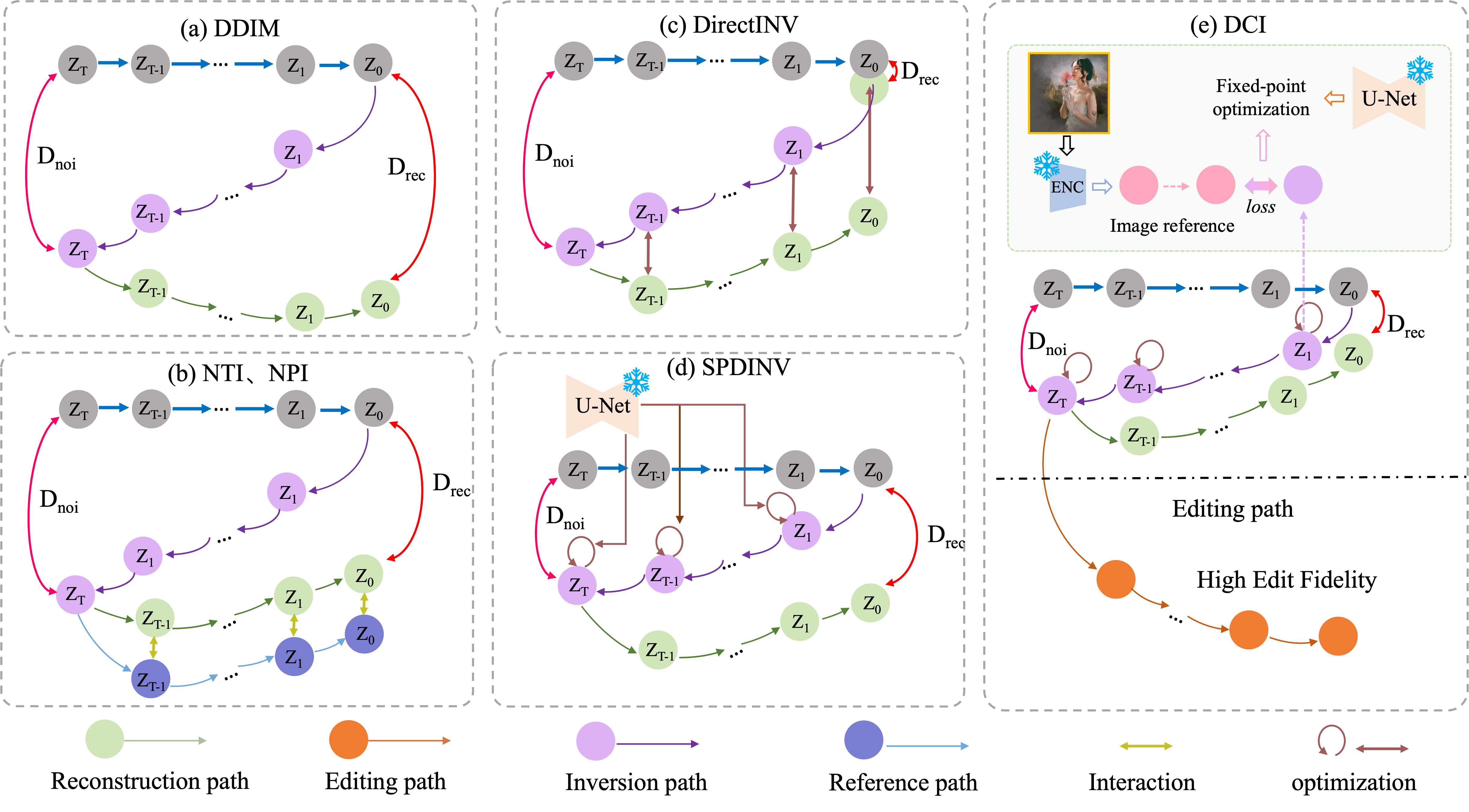}
    \caption{\textbf{Pipelines of different inversion methods in diffusion-based image editing. 
    }
Each sub-figure illustrates the specific process:
(a) DDIM inversion; (b) NTI and NPI; (c) DirectInv; (d) SPDInv; (e) our Dual-Conditional Inversion(DCI). Obviously, DCI significantly reduces both latent noise gap($D_{noi}$) and reconstruction error($D_{rec}$).
}
    \label{fig:all_method}
    \vspace{-5mm}
\end{figure}

Diffusion models have made significant progress in the field of generative artificial intelligence. Among them, latent Diffusion Models (LDMs)~\cite{LDM} perform the diffusion process in a compressed latent space rather than the pixel space, enabling more efficient and high-quality image generation and editing. 
This architectural design has made LDMs a powerful and flexible backbone for a wide range of downstream tasks, such as text-to-image generation~\cite{nichol2021glide,ramesh2022hierarchical,saharia2022photorealistic}, image editing~\cite{meng2021sdedit,masactrl,plugandplay,brooks2023instructpix2pix}, image restoration~\cite{lin2024diffbir,wang2024exploiting,wu2024one}, style transfer~\cite{wang2023stylediffusion,wang2024instantstyle,chung2024style}, \emph{etc}. In the image editing tasks, the editing is achieved by manipulating the diffusion latent representations. However, in most cases, the corresponding latent representation for a given image is not directly available, which means that we must first perform an inversion process to obtain their latent representations.

The earliest inversion method is DDPM~\cite{ho2020denoising}, and it has inspired the development of numerous related methods~\cite{tsaban2023ledits,Brack2023LEDITSLI,huberman2023edit}.
DDPMs add random noise at each timestep, which leads to the loss of information contained in the original image, resulting in poor reconstruction and editing effects. DDIM inversion~\cite{song2020denoising,dhariwal2021diffusion} reformulates the diffusion process to be deterministic as solving an implicit equation under the assumption that consecutive points along the denoising trajectory remain close. However, in practice, especially when using a limited number of denoising steps, this assumption often breaks down, leading to significant inaccuracies in the inversion results. In order to improve the reconstruction effect of DDIM inversion, 
multiple works have proposed effective optimization methods,
such as null-text embedding(NTI)~\cite{NTI} and negative prompt(NPI)~\cite{NPI} in the inversion process. As illustrated in figure~\ref{fig:all_method}, both NTI and NPI attempt to reduce the reconstruction gap($D_{rec}$) by optimizing the text embeddings. 
In the meanwhile, the researchers have
 proposed some alternative solutions from a non-optimization perspective. For instance, DirectInv~\cite{directinv} introduces a target-aware branch to correct the source branch trajectory, improving reconstruction quality. It performs well especially in terms of content preservation, and it is faster than optimization-based inversion methods.
Renoise~\cite{garibi2024renoise} is based on the linear assumption that the direction from \(z_t\) to \(z_{t + 1}\) can be approximated by the reverse direction from \(z_t\) to \(z_{t - 1}\). By calculating the direction from \(z_t\) to \(z_{t + 1}\) multiple times and taking the average, a more accurate direction from \(z_t\) to \(z_{t + 1}\) could be obtained. 
SPDInv~\cite{li2024source} SPDInv uses an optimization method to bridge the latent gap on each timestep, but the improvement
of reconstruction gap ($D_{rec}$) is limited. Although these methods have achieved certain success, they still face an intrinsic trade-off between reconstruction accuracy and editing flexibility. As illustrated in figure~\ref{fig:all_method}, such approaches struggle to reconcile semantic precision with structural consistency, particularly when textual supervision is sparse or ambiguous.

In this work, we present \textbf{Dual-Conditional Inversion (DCI)}, a new perspective on diffusion-based image editing that unifies text and image conditioned inversion within a fixed-point optimization framework. DCI addresses this limitation by introducing a dual-conditioning mechanism: it jointly leverages the source prompt \( p_s \) and the reference image \( x_0 \) to guide the inversion process. At the core of our formulation is a two-stage iterative procedure. The first stage, \textit{reference-guided noise correction}, refines the predicted noise at each timestep by anchoring it to a visually grounded reference derived from the source image. The second stage, \textit{fixed-point latent refinement}, imposes self-consistency by optimizing each latent variable \( z_t \) as a fixed point of the generative trajectory defined by DDIM dynamics. Formally, we cast inversion as a dual-conditioned fixed-point optimization problem that minimizes two objectives: (1) the discrepancy between the predicted and reference noise vectors across timesteps, and (2) the reconstruction error between the generated image and the original reference. This formulation not only improves inversion stability but also yields latent representations that are inherently editable and semantically aligned.

To sum up, our framework enables a plug-and-play integration with a variety of existing diffusion models, requiring neither retraining nor any modification to the original model. Through extensive experiments across multiple editing tasks, DCI achieves superior reconstruction quality and editing fidelity when compared to prior inversion baselines. Moreover, we demonstrate that the proposed dual-conditional fixed-point formulation facilitates stable convergence and generalizes well across a wide range of editing scenarios, highlighting the robustness and scalability of the proposed approach.

\section{Related Work}
\vspace{-0.1in}
\subsection{Image Editing with Diffusion Models}
In recent years, a large number of works based on diffusion models in the field of image editing demonstrate significant potential and adaptability across diverse tasks. 
These methods utilize diverse forms of guidance, such as text prompts, image references and segmentation maps to achieve editing objectives.~\cite{kwon2022diffusion,kim2022diffusionclip,couairon2022diffedit,huang2023region,li2025unsupervised} These advances better enable the ability to maintain editing precision and semantic consistency. The rapid development of diffusion models has significantly improved image generation capabilities. Among them, the widespread use of models such as GLIDE~\cite{nichol2021glide}, Imagen~\cite{saharia2022photorealistic}, DALL·E2~\cite{ramesh2022hierarchical}, and Stable Diffusion(SD)~\cite{LDM} has gradually expanded downstream tasks based on image generation. Prompt-to-Prompt(P2P)~\cite{hertz2022prompt} modifies cross-attention maps in diffusion models to enable text-driven image editing while preserving spatial structure through localized prompt adjustments. Pix2pix-zero~\cite{parmar2023zero} achieves zero-shot image-to-image translation by aligning latent features with text guidance. Plug-and-Play~\cite{plugandplay} integrates task-specific modules into pretrained diffusion backbones without retraining. MasaCtrl~\cite{masactrl} enhances real-time spatial control in diffusion models by injecting mask-guided attention constraints for precise region-specific manipulation. IP-Adapter~\cite{ye2023ip} injects visual features into the attention mechanism, enabling personalized generation without fine-tuning. ControlNet~\cite{zhang2023adding} introduces an auxiliary network to condition diffusion models on structural inputs like edges or poses. Some recent efforts have proposed different approaches to improve the precise of image editing from various perspectives~\cite{wei2023elite,mou2024diffeditor,ruiz2023dreambooth,bahjat2023}. Despite these methods have shown promising results, they often suffer from editing failures due to inversion methods. Our DCI improves upstream inversion to enhance downstream editing fidelity.
\vspace{-2.5mm}
\subsection{Inversion methods of diffusion models}
The earliest inversion methods include DDPM~\cite{huberman2023edit} and DDIM~\cite{DDIM}. DDPM generates high-quality images by progressively adding noise in a forward process and learning the reverse denoising process.~\cite{de2021diffusion,song2020score} Building on this foundation, DDIM introduces a deterministic sampling mechanism. Its near-invertible properties provide a crucial foundation for subsequent image inversion and editing techniques.
Researchers have conducted in-depth and extensive studies on the inversion process of diffusion models to achieve both efficiency and precision. Some methods focus on optimizing text embedding~\cite{NTI,NPI,Prox}. Null-Text Inversion (NTI)~\cite{NTI} adjusts latent encodings and text embeddings to reconstruct the original image. To improve efficiency, Negative-Prompt Inversion (NPI)~\cite{NPI} and its enhancements, including Proximal Guidance~\cite{Prox}, have emerged to reduce the reliance on time-consuming optimization processes. EDICT~\cite{wallace2023edict}, for example, achieves exact invertibility through coupling transformations, while methods like Direct Inversion~\cite{directinv} and Fixed-Point Inversion~\cite{meiri2023fixed} focus on simplifying the inversion process. The former decouples the diffusion branches, while the latter utilizes fixed-point iteration theory to ensure high reconstruction quality while reducing computational overhead.
Many inversion techniques also particularly focus on improving downstream editing tasks~\cite{li2024source,dong2023prompt}. For example, Source Prompt Disentangled Inversion (SPDInv)~\cite{li2024source} aims to decouple image content from the original text prompt, enhancing editing flexibility and accuracy. Specialized inversion and editing frameworks have been developed for specific editing needs~\cite{li2023stylediffusion,shi2024dragdiffusion}. Additionally, the concept of inversion has been extended to broader domains~\cite{gal2022image,huang2024reversion,dong2023prompt,cho2024noise,zhang2023real}. Textual Inversion proposes learning new text embeddings to represent user-specific concepts for personalized image generation~\cite{gal2022image}. ReVersion~\cite{huang2024reversion} further explores learning and inverting relational concepts from images. Meanwhile, works like Aligning Diffusion Inversion Chain~\cite{zhang2023real} focus on generating high-quality image variants by aligning inversion chains.

Although the above methods have solved the reconstruction problem to a certain extent, they may bring artifacts and inconsistent details when applied to editing tasks. Most of the time, they only focus on the text prompt or the original image, but do not integrate them. In our work, we propose a simple but effective method to fuse the text prompt and source image in the form of fixed-point iteration. Our method improves the editing fidelity a lot and shows inspiring results.

\section{Dual-Conditional Inversion}
\subsection{Motivation and Problem Formulation}

\begin{figure}
    \centering
    \includegraphics[width=1\linewidth]{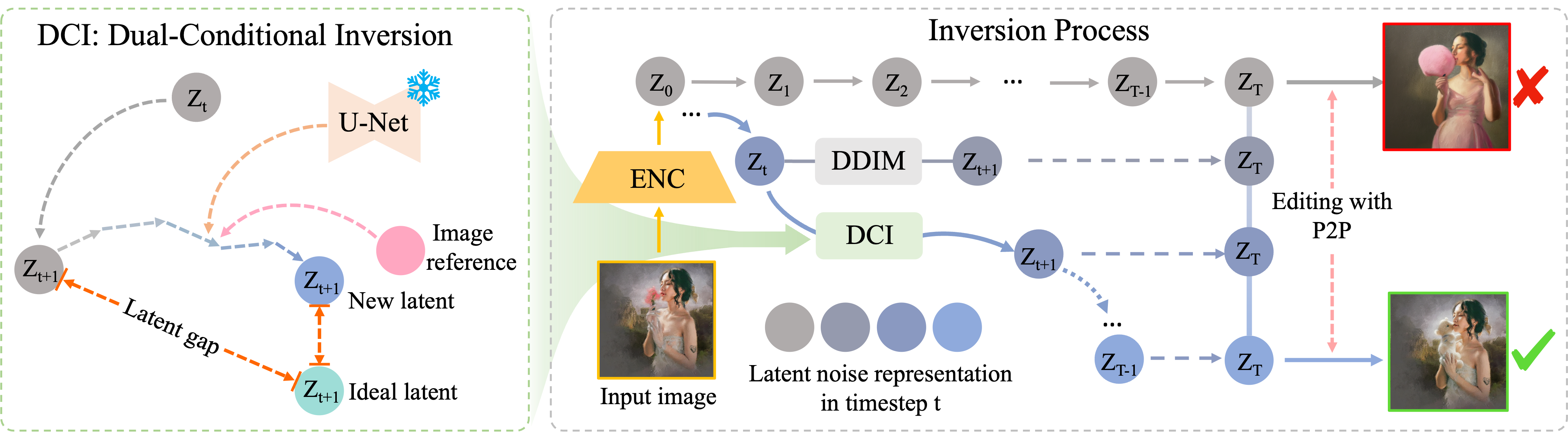}
    \caption{\textbf{Inversion process of DCI}. The green box on the left illustrates DCI, which use dual-conditional guidance to reduce the latent gap. The right describes how DCI modifies the inversion process and generate the latent noise code. It also shows our method can improve the editing method.}
    \label{fig:method_fig}
\end{figure}
In most diffusion-based image editing frameworks, the inversion process plays a foundational role: it converts an image to the latent noise representation from which the image can be reconstructed and edited. However, diffusion models inherently lack an explicit and exact inverse process to convert an image back to its corresponding latent noise representation. Ideally, a successful inversion would yield a latent code $z_T$  that faithfully preserves both the semantic content and structural details of the input image, thereby enabling accurate reconstruction and precise downstream editing. However, the information loss caused by repeated noise injection in inversion process makes perfect inversion unattainable, even when auxiliary constraints such as text prompts or reference images are employed.

To analyze the limitations of current inversion strategies, we begin with DDIM (Denoising Diffusion Implicit Models)~\cite{DDIM}, a deterministic variant of DDPM~\cite{ho2020denoising}. DDIM defines a closed-form sampling process that generates a latent image $z_0$ from Gaussian noise $z_T \sim \mathcal{N}(0, \mathbf{I})$ as follows:
\begin{equation}
    z_{t-1}
= \frac{\sqrt{\alpha_{t-1}}}{\sqrt{\alpha_t}} z_t + \sqrt{\alpha_{t-1}} \left( \sqrt{\frac{1}{\alpha_{t-1}} - 1} - \sqrt{ \frac{1}{\alpha_t} -1} \right) \epsilon_{\theta}(z_t, t, c),
\end{equation}
Where $\alpha_t$ denotes the cumulative noise schedule, and $\epsilon_{\theta}$ represents the noise predicted by a U-Net, conditioned on the current timestep $t$ and a control input $c$ (e.g. , a text prompt). However, using only a text prompt as $c$ is insufficient for accurately reconstructing the original image. Recent methods such as ControlNet~\cite{zhang2023adding} and IP-Adapter~\cite{ye2023ip} enrich the conditioning input $c$ with visual features from the original image, thereby improving generation quality. Nevertheless, these methods are often computationally expensive and difficult to integrate into the inversion process. Ideally, inversion requires recovering $z_t$ from a known $z_{t-1}$, which leads to the following “ideal inversion” formula:
\begin{equation}
z_t = C_{t,1} \cdot z_{t-1} + C_{t,2} \cdot \epsilon_{\theta}(z_t, t, c_{\text{ideal}}),
\end{equation}
where the coefficients are defined as: $C_{t,1} = \frac{ \sqrt{\alpha_t} }{ \sqrt{\alpha_{t-1}} }, \quad C_{t,2} = \sqrt{\alpha_t} \left( \sqrt{ \frac{1}{\alpha_t} - 1 } - \sqrt{ \frac{1}{\alpha_{t-1}} - 1 } \right).$

However, in practice, this expected inversion is not feasible because the ideal latent  $z_t$  is not available when performing the inversion step from  $z_{t-1}$. Thus, the DDIM inversion process approximates this update by feeding  $(z_{t-1}, t-1, c)$  into the inversion process instead of  $(z_t, t, c)$, leading to the practical inversion formula:
\begin{equation}
    z_t = C_{t,1} \cdot z_{t-1} + C_{t,2} \cdot \epsilon_{\theta}(z_{t-1}, t-1, c).
\label{get_zt}
\end{equation}

This approximation breaks the strict reversibility of the ODE-based formulation and introduces temporal mismatch error between the predicted noise and the actual generative trajectory. Since the diffusion model assumes infinitesimal step size for reversibility (akin to a continuous ODE), using coarse discrete steps and mismatched inputs (i.e., $\epsilon_\theta(z_{t-1}, t-1, c)$ instead of the ideal $\epsilon_\theta(z_t, t, c))$ induces systematic error at each timestep.

If a real image and its corresponding text prompt are given, the image generated directly using the text prompt will be very different from the real image. The reason arises from the inaccuracy of text prompt and randomness in the generation process. From this perspective, there are also errors in the use of $\epsilon_\theta(z_t, t, c))$  for the inversion process. This error is also accumulated over time, resulting in the final $z_t$ not being well applied to reconstruction and editing. In previous work, SPDInv~\cite{li2024source} transforms the inversion process into a search problem that satisfies fixed-point constraints. The pre-trained diffusion model is used to make the inversion process as independent of the source prompt as possible, thereby reducing the gap between $\epsilon_\theta(z_t, t, c))$  and $\epsilon_\theta(z_{t-1}, t-1, c)$. Although SPDInv narrows the gap between $\epsilon_\theta(z_t, t, c))$ and $\epsilon_\theta(z_{t-1}, t-1, c)$. However, in the previous analysis, $\epsilon_\theta(z_t, t, c))$ is not an ideal noise. The ideal noise should not only be separated from the source prompt, but also retain more information of the original image. What needs to be reduced is the difference between $\epsilon_\theta(z_t, t, c_{ideal})$ and $\epsilon_\theta(z_{t-1}, t-1, c)$, and this difference will appear in each inversion process and accumulate in the final output.

To achieve high-fidelity inversion, it is essential to minimize the discrepancy between the predicted noise and the ideal generative direction at each timestep. This requires not only disentangling the inversion process from the source prompt(mentioned in~\cite{li2024source}), but also preserving as much information from the original image as possible. Addressing both aspects simultaneously is key to reducing cumulative errors and improving the reconstruction and editability of the inverted latent noise representations in diffusion-based image editing.

\subsection{Dual-Conditional Inversion (DCI)}

To address the limitations of existing inversion methods, we propose Dual-Conditional Inversion (DCI), a novel framework that enhances the latent noise representations in diffusion models. DCI leverages both the original image and text prompt to guide the inversion process, ensuring high-fidelity reconstruction and improved editability. Unlike prior approaches, DCI integrates these into a dual-conditional fixed-point optimization pipeline. The method consists of two key stages: \textit{reference-guided noise correction} that anchors the inversion to the source image, and \textit{fixed-point latent refinement} that ensures self-consistency with the generative process.

\subsubsection{Reference-Guided Noise Correction}
The first stage of DCI introduces a reference-based constraint to align the predicted noise with the source image. At each DDIM timestep \( t \), we compute an initial noise estimate conditioned on the source prompt \( p_s \):
\begin{equation}
\hat{\epsilon}_{\text{raw}} = \epsilon_\theta(z_t, t, p_s).
\label{e_raw}
\end{equation}
 where \( \epsilon_\theta \) is the noise prediction model (e.g., a U-Net) and \( z_t \) is the current latent. However, \( \hat{\epsilon}_{\text{raw}} \) often deviates from the ideal noise due to the coarse constraint of \( p_s \). While this prediction reflects prompt-level semantics, it often deviates from the actual noise corresponding to the input image due to limited grounding provided by textual information alone. To address this, we introduce a visual reference signal by extracting a reference noise vector \( \epsilon_{\text{ref}} \) from the source image latent \( z_0 \), which is obtained via a pretrained VAE encoder $E$. The reference noise is defined as:
\begin{equation}
\epsilon_{\text{ref}} = E(z_0).
\end{equation}

 The $\epsilon_{\text{ref}}$ serves as an anchor to guide the correction of prompt-based noise estimation. To enforce alignment between the prompt-predicted noise and the image-derived reference, we define a reference alignment loss:
\begin{equation}
\mathcal{L}_{\text{ref}} = \left\| \hat{\epsilon}_{\text{raw}} - \epsilon_{\text{ref}} \right\|_2 .
\label{Lref}
\end{equation}

Equation~\ref{Lref} penalizes the discrepancy between the two noise vectors. A one-step gradient-based correction is then applied to refine the noise prediction:
\begin{equation}
\hat{\epsilon} = \hat{\epsilon}_{\text{raw}} - \lambda \cdot \nabla_{\hat{\epsilon}_{\text{raw}}} \mathcal{L}_{\text{ref}} .
\end{equation}
where \( \lambda \) is a hyperparameter that controls the correction strength. This update adjusts the predicted noise in a direction that reduces its divergence from the reference signal, effectively grounding the inversion in visual structure. As a result, this correction improves reconstruction fidelity and ensures that the denoising trajectory remains semantically and perceptually consistent with the original image, particularly in scenarios where the prompt is ambiguous or underspecified.

\subsubsection{Fixed-Point Latent Refinement}
After correcting the noise estimate, we proceed to update the latent variable \( z_t \) using the DDIM inversion formula.  This step changes the inversion trajectory from timestep \( t-1 \) to \( t \), based on the corrected noise \( \hat{\epsilon} \):
\begin{equation}
z_t = C_{t,1} \cdot z_{t-1} + C_{t,2} \cdot \hat{\epsilon},
\end{equation}
where \( C_{t,1} = \frac{\sqrt{\alpha_t}}{\sqrt{\alpha_{t-1}}} \) and \( C_{t,2} = \sqrt{\alpha_t} \left( \sqrt{\frac{1}{\alpha_t} - 1} - \sqrt{\frac{1}{\alpha_{t-1}} - 1} \right) \), and \( \alpha_t \) is the noise schedule. While this deterministic update follows the DDIM trajectory, it remains sensitive to error accumulation during the inversion process. As such, it may introduce perturbations into the latent dynamics, ultimately affecting reconstruction and editing fidelity. To improve stability and enforce consistency with the forward generative process, DCI introduces a fixed-point refinement step that iteratively corrects the latent by treating it as a fixed-point problem of the DDIM inversion at each timestep. Specifically, we define the latent update function:

\begin{equation}
f_\theta(z_t) = C_{t,1} \cdot z_{t-1} + C_{t,2} \cdot \epsilon_\theta(z_t, t, p_s).
\end{equation}
The objective is to find a latent \( z_t \) such that:
\begin{equation}
z_{t} = f_\theta(z_t).
\end{equation}

To achieve this, we minimize the following fixed-point self-consistency loss:
\begin{equation}
\mathcal{L}_{\text{fix}} = \left\| f_{\theta}(z_t) - z_t \right\|_2
\end{equation}

We iteratively refine \( z_t \) using gradient descent:
\begin{equation}
z_t = z_t - \eta \cdot \nabla_{z_t} \mathcal{L}_{\text{fix}},
\end{equation}
where \( \eta \) is the learning rate of refinement process. This fixed-point update step is repeated for up to \( K \) iterations or until the convergence criterion \( \mathcal{L}_{\text{fix}} < \delta \) is satisfied. In practice, our method converges rapidly within a few iterations(usually no more than 10 iterations), which ensures computational efficiency without compromising reconstruction quality. By explicitly enforcing this self-consistency constraint, DCI stabilizes the inversion trajectory and reduces artifacts that arise from misaligned latents. This refinement step not only enhances reconstruction quality but also improves the reliability and flexibility of downstream editing operations.

\begin{algorithm}
	\renewcommand{\algorithmicrequire}{\textbf{Input:}}
	\renewcommand{\algorithmicensure}{\textbf{Output:}}
	\caption{Dual-Conditional Inversion (DCI)}
	\label{alg:dci}
	\begin{algorithmic}[1]
		\REQUIRE Source image latent $z_0$, DDIM steps $T$, source prompt $p_s$, maximal optimization rounds $K$, threshold $\delta$, image guidance strength $\lambda$, fixed-point learning rate $\eta$, reference noise $\epsilon_{\text{ref}}$
		\ENSURE Inversion noise $z_T$
		\FOR{$t = 1$ to $T$}
			\FOR{$i = 1$ to $K$}
				\STATE Get $z_t$ from $z_{t-1}$ based on \eqref{get_zt}
				\STATE Predict noise $\hat{\epsilon}_{\text{raw}}$ based on \eqref{e_raw}
				\STATE Compute $\mathcal{L}_{\text{ref}} = \left\| \hat{\epsilon}_{\text{raw}} - \epsilon_{\text{ref}} \right\|_2$
				\STATE Apply correction: $\hat{\epsilon} = \hat{\epsilon}_{\text{raw}} - \lambda \cdot \nabla_{\hat{\epsilon}_{\text{raw}}} \mathcal{L}_{\text{ref}}$
				\STATE \textbf{Update} $z_t$ using $\hat{\epsilon}$
				\STATE Calculate $\mathcal{L}_{\text{fix}} = \left\| f_{\theta}(z_t) - z_t \right\|_2$
				\STATE Update $z_t = z_t - \eta \cdot \nabla_{z_t} \mathcal{L}_{\text{fix}}$
				\STATE \textbf{if} $\mathcal{L}_{\text{fix}} < \delta$ \textit{then break} \textbf{end if}
			\ENDFOR
		\ENDFOR
	\end{algorithmic}
\end{algorithm}

\subsubsection{Algorithm Summary}
The complete Dual-Conditional Inversion (DCI) process is summarized in Algorithm~\ref{alg:dci}. At each DDIM timestep, DCI first performs \textit{Reference-Guided Noise Correction} to obtain a visually grounded noise estimate \( \hat{\epsilon} \) by combining prompt-based prediction and reference-derived supervision. Then it is followed by \textit{Fixed-Point Latent Refinement}, which iteratively updates the latent \( z_t \) to satisfy a self-consistency condition defined by the DDIM inversion dynamics. The dual conditioning on both the source prompt \( p_s \) and the reference image (via \( \epsilon_{\text{ref}} \)) ensures that the final inverted latent \( z_T \) closely approximates the ideal generative noise \( z_T^* \), which leads to reliable reconstruction and high-fidelity, better structure-preserving editing.

\begin{figure}[!t]
    \centering
    \includegraphics[width=1\linewidth]{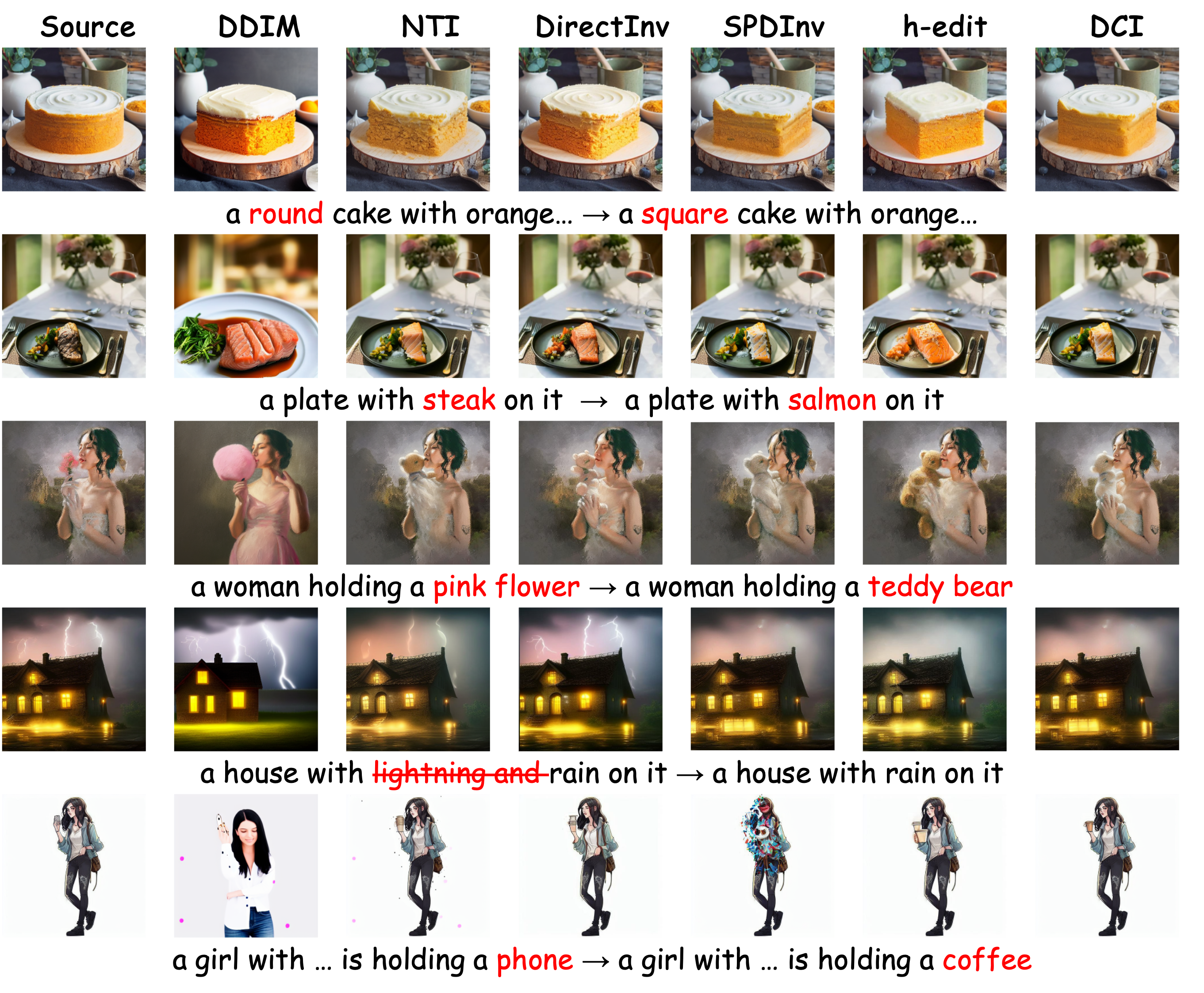}
    \vspace{-0.1in}
    \caption{\textbf{Visual results of different inversion methods with P2P on PIE-Bench.} Each method is identified at the top of its respective column, while detailed editing information appears beneath each corresponding row. DCI(ours) demonstrates significant enhancements over existing methods.}
    \label{fig:comparison}
    \vspace{-3mm}
\end{figure}
\section{Experiments}
We conduct extensive experiments to evaluate the effectiveness of Dual-Conditional Inversion (DCI). This section is organized as follows. In Section~\ref{sec:setup}, we introduce the datasets, evaluation metrics and experimental settings. Section~\ref{sec:comparison} compares DCI with representative inversion methods across multiple aspects quantitatively and qualitatively. In Section~\ref{sec:gap}, we investigate how DCI reduces both the latent noise gap and the reconstruction error. Finally, Section~\ref{sec:ablation} presents an ablation study to assess the impact of key hyperparameters and design choices.
\subsection{Experimental Setups}
\label{sec:setup}
\textbf{Evaluation Metrics.} 
We mainly use DINO score~\cite{caron2021emerging}, Peak Signal-to-Noise Ratio (PSNR), Mean Squared Error (MSE), Structural Similarity Index (SSIM), and Learned Perceptual Image Patch Similarity (LPIPS)~\cite{zhang2018unreasonable}
to evaluate the performance of DCI from multiple perspectives. 
We use the DINO score to evaluate the overall structural similarity of the generated images, while the CLIP score~\cite{radford2021learning} is employed to quantify the alignment between the generated image and the given prompt. For background preservation and image fidelity, we report PSNR, MSE, SSIM, and LPIPS, with all metrics computed specifically over the annotated regions in the dataset from DirecInv~\cite{directinv}.
Both the DINO and CLIP scores are calculated over the entire image to capture global consistency, whereas the remaining metrics focus on local quality within specified regions.

\textbf{Datasets.}
We verifies the effectiveness of our proposed DCI method mainly on the PIE-Bench~\cite{directinv}, which comprises $700$ images featuring $10$ distinct editing types. It provides five annotations on each image: source image prompt, target image
prompt, editing instruction, main editing body, and the editing mask. The calculation of region-specific metrics heavily relies on the editing mask, as the editing is expected to occur only within the annotated region. We also use the \textit{COCO2017}~\cite{lin2014microsoft} to test the application of our method in a wider range of scenarios.

\textbf{Other Settings.} In our experiments, we utilize Stable Diffusion v1.4 as the base model with DDIM sampling steps of 50 and a Classifier-Free Guidance (CFG) scale of 7.5. These settings are the same as those used in the baselines. For DCI, we set the hyper-parameters to \( K=5 \), \( \lambda=2 \), and \( \eta=0.001 \). All experiments and validations are conducted on a single NVIDIA RTX 4090 GPU.
\subsection{Comparisons with Inversion-Based Editing Methods}
\label{sec:comparison}
\begin{table}[t]
\caption{Performance comparison of inversion-based methods under the Prompt-to-Prompt (P2P) editing engine~\cite{dong2023prompt} on PIE-Bench. Metrics include DINO ($\downarrow$), PSNR ($\uparrow$), LPIPS ($\downarrow$), MSE ($\downarrow$), SSIM ($\uparrow$), and CLIP ($\uparrow$). Best and second-best results are highlighted in \redContent{red} and \blueContent{blue}, respectively. DCI (ours) achieves the best performance across all metrics.}
\vspace{0.1in}

    \label{tab:metrice_edit}
    \centering
    \begin{tabular}{l|c|c|c|c|c|c|c}
   \toprule
   \makecell[c]{Inversion} & \makecell[c]{Editing Engine} & \makecell[c]{DINO$\downarrow$ \\ $\times10^3$} & PSNR$\uparrow$ & \makecell[c]{LPIPS$\downarrow$ \\ $\times10^3$} & \makecell[c]{MSE$\downarrow$ \\ $\times10^4$} & \makecell[c]{SSIM$\uparrow$ \\ $\times10^2$} & \makecell[c]{CLIP}$\uparrow$ \\
   \midrule
DDIM~\cite{DDIM}& P2P & 69.43 & 17.87 & 208.80 & 219.88 & 71.14 & 25.01   \\
NTI~\cite{NTI}& P2P & 13.44 & 27.03 & 60.67 & 35.86 & 84.11 & 24.75  \\
NPI~\cite{NPI}& P2P & 16.17 & 26.21 & 69.01 & 39.73 & 83.40 & 24.61  \\
AIDI~\cite{AIDI}& P2P & 12.16 & 27.01 & 56.39 & 36.90 & 84.27 & 24.92 \\
NMG~\cite{cho2024noise}
& P2P & 23.50 & 25.83 & 81.58 & 107.95 & 82.31 & 24.05  \\
DirectINV~\cite{directinv} & P2P & 11.65 & 27.22 & 54.55 & 32.86 & 84.76  & 25.02 \\
ProxEdit~\cite{Prox} & P2P & 11.87 & 27.12 & 45.70 & 32.16 & 84.80 & 24.28 \\
SPDInv~\cite{li2024source}&P2P & \blueContent{8.81} & \blueContent{28.60} & \blueContent{36.01} & \blueContent{24.54} & \blueContent{86.23} & \blueContent{25.26} \\

\textit{h}-Edit~\cite{nguyen2025hedit} & P2P & 11.17 & 27.87 & 48.50 & 85.40 & 84.80 & 25.30 \\
\midrule

DCI(ours)&P2P & \redContent{6.07} & \redContent{29.38} & \redContent{33.01} & \redContent{21.28} & \redContent{87.14} & \redContent{25.52}   \\

   \bottomrule
\end{tabular}
\vspace{-2mm}

\end{table}

We compare DCI with several inversion-based methods quantitatively and qualitatively. These methods includes DDIM inversion~\cite{DDIM}, Null-text inversion (NTI)~\cite{NTI}, Negative prompt inversion (NPI)~\cite{NPI}, AIDI~\cite{AIDI}, Noise Map Guidance (NMG)~\cite{cho2024noise}, Direct Inversion (DirectINV)~\cite{directinv}, ProxEdit~\cite{Prox}, SPDInv~\cite{li2024source} and \textit{h}-Edit~\cite{nguyen2025hedit}. We mainly evaluate under the Prompt-to-Prompt (P2P) editing engine on PIE-Bench. 
As Table \ref{tab:metrice_edit} shows, DCI (ours) outperforms all methods across DINO, PSNR, LPIPS, MSE, SSIM, and CLIP metrics. Compared to the second-best method, SPDInv, DCI achieves significant improvements, including a 31.1\% reduction in DINO (6.07vs.8.81), 8.3\% reduction in LPIPS (33.01vs.36.01), and 13.3\% reduction in MSE (21.28vs.24.54). At the same time, It is also higher than SPDInv in other metrics(PSNR,SSIM,CLIP). Compared with other methods listed in Table \ref{tab:metrice_edit}, our method has a greater improvement. These results underscore DCI's superior accuracy and robustness for high-fidelity image editing.

Figure \cref{fig:comparison} presents a visual comparison with the P2P engine. The first row presents cake images frequently used for comparative analysis in existing methods. Most approaches show satisfactory results. In contrast, the second row demonstrates that our method enhances detail representation in salmon. The third row illustrates when modifying features such as hands or mouth, previous methods will fail. However, our DCI achieves this task while maintaining high-quality output. In the fourth row, our method achieves better background color fidelity and reduces lighting artifacts compared to others. The fifth row highlights our method's robust performance in local part editing while preserving overall consistency across other image regions.

Due to the page limit, we provide more visual and quantitative results under different editing engines (such as masactrl) in the \textbf{supplementary material}. We can draw similar conclusions to the above from these experimental results.

\vspace{-1mm}
\subsection{Reduction of Noise and Reconstruction Gap by DCI}
\label{sec:gap}

We conduct experiments and confirm that our method can reduce the gap between noise and reconstruction ($D_{noi}$ and $D_{rec}$ as depicted in Figure~\ref{fig:all_method}). We randomly select 100 captions from the PIE-Bench and use Stable Diffusion V1.4 to generate images.
\begin{wrapfigure}{r}{0.45\textwidth}
    \centering
    \includegraphics[width=0.35\textwidth]{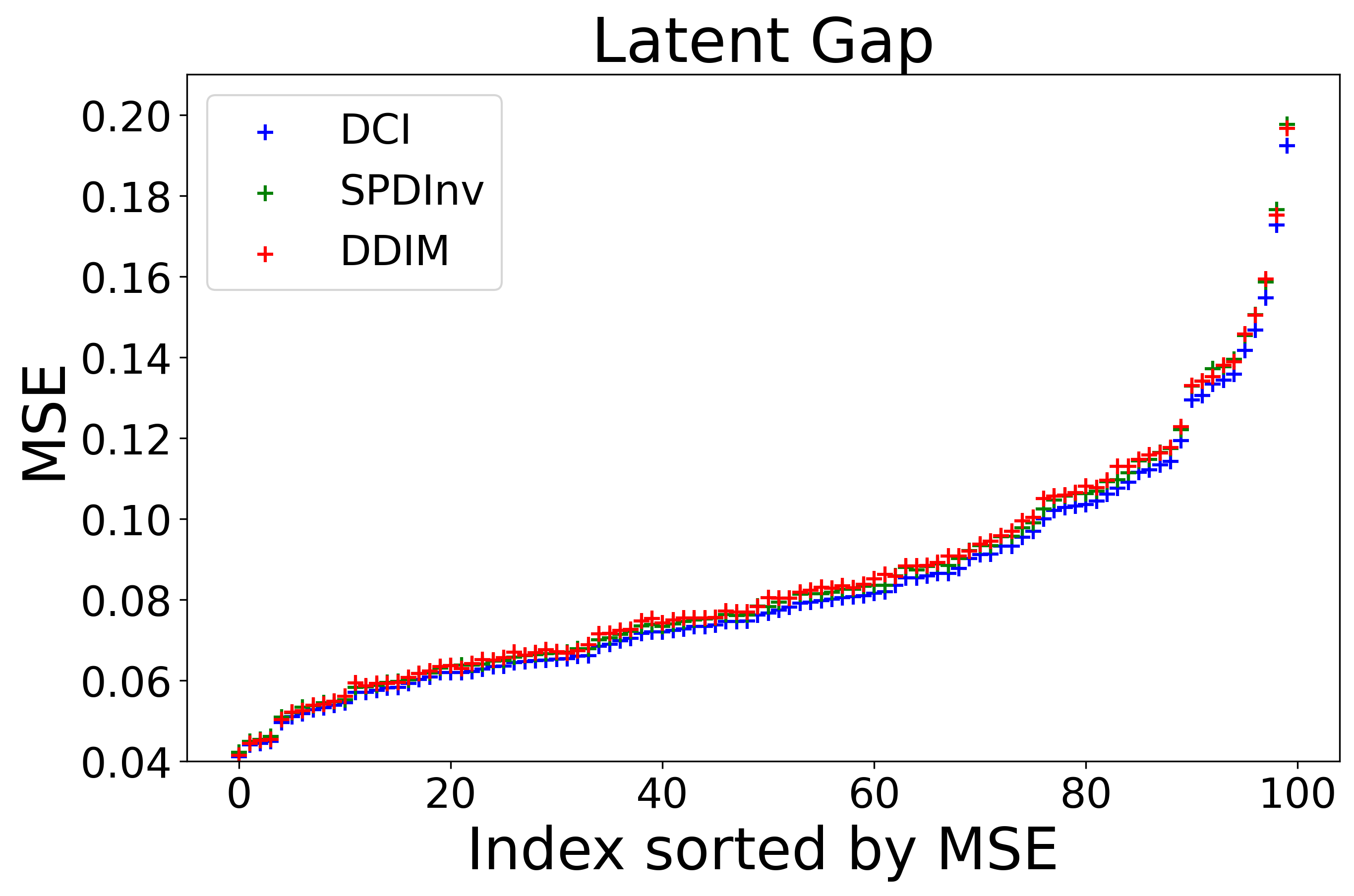} %
    \vspace{-1mm}
    \caption{\small Illustration of our framework.}
    
    \label{fig:latent_gap}
\end{wrapfigure}
We initialize $z_T$ with a fixed random seed, treating it as the ideal noise input for every image at the initial timestep of the diffusion process. The final generated image serves as a reference for reconstruction accuracy assessment. We visualize and evaluate the performance of our method with DDIM~\cite{DDIM} and SPDInv~\cite{li2024source}.

For latent gap analysis, we visualize the $z_T$ gap obtained by these methods in figure~\ref{fig:latent_gap}. The concentration of the data shows that our method is closer to the ideal noise. For reconstruction gap evaluation, we use both MSE and CLIP scores.
DDIM yields an MSE of $1.32 \times 10^{-4}$, SPDInv achieves $1.21 \times 10^{-4}$, while DCI obtains the lowest error at $1.12 \times 10^{-4}$. The CLIP Scores are $26.91$ for DDIM, $26.92$ for SPDInv, and $26.94$ for DCI.
Comparatively, our technique demonstrates superior performance over DDIM and SPDInv based on these metrics.

\begin{table*}[!t]
\vspace{-5mm}
\caption{Ablation study on the hyper-parameters of DCI with PIE-Bench.}
\vspace{0.08in}
\label{tab:Ablation}
\centering
\renewcommand{\arraystretch}{1.2}
\rowcolors{2}{gray!15}{white}
\resizebox{\textwidth}{!}{%
    \begin{tabular}{l|c|c|c|c|c|c}
    \toprule
    \textbf{Hyper-parameter} & \textbf{$\text{DINO}_{\times10^3}\downarrow$} & \textbf{PSNR$\uparrow$} & \textbf{$\text{LPIPS}_{\times10^3}\downarrow$} & \textbf{$\text{MSE}_{\times10^4}\downarrow$} & \textbf{$\text{SSIM}_{\times10^2}\uparrow$} & \textbf{CLIP$\uparrow$} \\
    \midrule

    $K=2$      &6.13 & 29.32 & 33.10 & 21.50 & 87.11 & 25.49 \\
    $K=5$     & 6.07  & 29.38 & 33.01  & 21.28 & 87.14 &25.52\\
    $K=10$    &6.17 & 29.29 & 33.17 & 21.56 & 87.12 & 25.51  \\
    \midrule
    $\lambda=1$    & 6.19  & 29.29 & 33.12  & 21.62 & 87.12 & 25.53 \\
    $\lambda=2$    & 6.07  & 29.38 & 33.01  & 21.28 & 87.14 & 25.52 \\
    $\lambda=5$    & 9.29  & 28.25 & 41.05  & 26.18 & 86.26 & 25.38 \\
    \midrule
    $\eta=0.0001$ & 6.72  & 28.80 & 35.93  & 23.72 & 86.70 & 25.50 \\
    $\eta=0.001$   & 6.07  & 29.38 & 33.01  & 21.28 & 87.14 & 25.52 \\
    $\eta=0.01$   & 35.29 & 23.05 & 88.90  & 83.84 & 81.18 & 25.02 \\
    \midrule
    \textbf{Default} & \textbf{6.07} & \textbf{29.38} & \textbf{33.01} & \textbf{21.28} & \textbf{87.14} & \textbf{25.52} \\
    \bottomrule
    \end{tabular}%
}
\end{table*}
\vspace{-1mm}
\subsection{Ablation Study}
\label{sec:ablation}
\vspace{-1mm}
Table~\ref{tab:Ablation} presents an ablation study on three key hyper-parameters of DCI: the number of optimization rounds ($K \in \{2,5,10\}$), the reference-guided noise correction weight ($\lambda  \in \{1,2,5\}$), and the learning rate ($\eta \in \{0.0001,0.001,0.01\}$). The method converges quickly, as even a small number of rounds ($K=2$) shows competitive results, and performance saturates by $K=5$. 
$\lambda=2$ achieves the best trade-off, while higher values such as $\lambda=5$ lead to significant degradation across all metrics, indicating over-dependence on inversion constraints. The learning rate $\eta=0.001$ provides the most stable and effective optimization; both smaller and larger values reduce reconstruction quality, with $\eta=0.01$ causing severe performance collapse. These results support the choice of the default configuration ($K=5$, $\lambda=2$, $\eta=0.001$) as optimal for balancing fidelity and stability.
\vspace{-1mm}

\section{Conclusion}
\vspace{-1mm}
In this paper, we introduce Dual-Conditional Inversion (DCI), a novel method that combines both the source prompt and the reference image to guide the inversion process. By formulating inversion as a dual-conditioned fixed-point optimization problem, DCI reduces both latent noise gap and reconstruction errors in diffusion models. 
Notably, DCI exhibits strong plug-and-play capability: it can be seamlessly integrated into existing diffusion-based editing pipelines without requiring model retraining or architecture modification. 
Extensive experiments demonstrate that our method achieves superior edit quality on benchmark datasets. Overall, DCI provides a robust, flexible, and easily deployable foundation for future research in diffusion-based tasks.

\bibliographystyle{plain}
\bibliography{ref}
\end{document}